# Empirical Analysis of Oral and Nasal Vowels of Konkani

Swapnil Fadte*, Edna Vaz†, Atul Kr. Ojha††, Ramdas Karmali*, Jyoti D. Pawar*

* Discipline of Computer Science & Technology, Goa Business School, Goa University
swapnil.fadte@unigoa.ac.in, rnk@unigoa.ac.in, jdp@unigoa.ac.in
†Govt. College of Arts Science and Commerce Quepem, Goa. And Dept. of Linguistics, University of Mumbai
edna.vaz22@gmail.com
††Insight SFI Research Centre for Data Analytics, Data Science Institute, University of Galway
atulkumar.ojha@insight-centre.org

**Abstract**

Konkani is a highly nasalised language which makes it unique among Indo-Aryan languages. This work investigates the acoustic-phonetic properties of Konkani oral and nasal vowels. For this study, speech samples from six speakers (3 male and 3 female) were collected. A total of 74 unique sentences were used as a part of the recording script, 37 each for oral and nasal vowels, respectively. The final data set consisted of 1135 vowel phonemes. A comparative F1-F2 plot of Konkani oral and nasal vowels is presented with an experimental result and formant analysis. The average F1, F2 and F3 values are also reported for the first time through experimentation for all nasal and oral vowels. This study can be helpful for the linguistic research on vowels and speech synthesis systems specific to the Konkani language.

**Keywords:** Konkani, oral vowel, nasal vowel, formant analysis

## 1. Introduction

Konkani belongs to the Indo-Aryan branch of the Indo-European family of languages. It is a member of the southern group of Indo-Aryan languages, and is most closely related to Marathi within this group (Miranda, 2018)). Konkani is mainly spoken in Goa and in some parts of the neighbouring states of Maharashtra, Karnataka, and Kerala, where Konkani speakers migrated after the Portuguese arrival in Goa. The 1991 Census of India records the number of Konkani speakers to be 1,760,607 out of which 602,626 (34.2 %) were from Goa, 312,618 (17.8 %) were from Maharashtra, 706,397 (40.1 %) from Karnataka, and 64,008 (3.6 %) from Kerala (Miranda, 2018). Konkani is written in different scripts in the regions where it is spoken.

### 1.1. Phonological Features of Konkani

As regards the phonology of Konkani, different scholars have mentioned different numbers of vowels and consonants in the language. Also, there is no consensus on the exact specification of the vowels as regards their place in the vocal tract. Nasalisation is phonemic in Konkani (as shown by i and ii below).

i  hɛ            tat͡ʃɛ           bʰurgɛ
   these.mas.pl.  he.gen.mas.pl.  children.mas.pl.
   'These are his (male) children.'

ii hɛ̃           tat͡ʃɛ̃          bʰurgɛ̃
   this.neut.sg. he.gen.neut.sg. child.neut.sg.
   'This is his child.'

### 1.2. Related Work on Konkani

(Sardesai, 1986), (Sardesai, 1993), in her dialect-specific work refers to nine Konkani Vowel phonemes: Front [i, e, ɛ]; Central [ɨ ə, a]; and Back [u, o, ɔ]. The author mentions that all these vowel phonemes can be nasalised. Her work is summarised in Table 1.

(Almeida, 1989) makes a reference to eight oral vowels for Konkani: Front [i, e, ɛ] and Back [u, o, ɔ, θ, a]. His classification of oral vowels is provided in Table 1. The author mentions that all vowels present in the language can be nasalised. Examples for both oral and nasal vowels are presented by him in his work. The author seems to consider vowel length to be phonemic in Konkani, which is not the case, at least for the Konkani varieties spoken in Goa.(Miranda, 2018) mentions nine Vowel phonemes for Konkani, along with their corresponding nasal counterparts. These are: [i, e, ɛ, ə, ʌ, a, u, o, ɔ].

(Fadte et al., 2022) provide a vowel chart for Konkani based on their acoustic analysis of vowels. They also provide the properties of vowel pairs which have different phonetic realisations but the same written representation in the script. Their vowel classification work is presented in Table 2, which includes equivalent vowels in different scripts, namely Devanagari, Roman and Kannada. Their work also acknowledges that all oral vowels could be nasalised.

| Author and Year | Vowels | Classification |
|---|---|---|
| (Sardesai, 1986) | i, e, ɛ, u, o, ɔ, a, ə, ɨ | Dialect-specific |
| (Almeida, 1989) | i, e, ɛ, u, o, ɔ, a, θ | General |
| (Miranda, 2018) | i, e, ɛ, u, o, ɔ, a, ə ʌ | General |
| (Fadte et al., 2022) | i, e, ɛ, u, o, ɔ, a, ə, ɨ | General |

Table 1: Comparison of Konkani Vowel classifications

### 1.3. Related Work on Other Languages

(Shosted et al., 2012) have presented work on Hindi nasal vowels, where F1-F2 values are used for calculating the

| Approximate IPA Notation | Equivalent grapheme | | | Examples | English meaning | Vowel type |
|---|---|---|---|---|---|---|
| | Roman | Devanagari | Kannada | | | |
| i | i | इ and ई | ಇ and ಈ | [ v iː s ] | twenty | Front |
| | | | | [ k ə v iː ] | poet | |
| e | e | ए | ಎ | [ p eː r ] | guava tree | |
| | | | | [ kʰ eː l ] | game | |
| | | | | [ m eː dʒ ] | count (imperative.2 p.sg.) | |
| ɛ | | ऍ | ಎ | [ p ɛː r ] | guava fruit | |
| | | | | [ kʰ ɛː l ] | play (imperative.2 p.sg.) | |
| | | | | [ m ɛː dʒ ] | table | |
| ɨ | | अ | ಅ | [ k ɨ ːr ] | do (imperative.2 p.sg.) | Central |
| | | | | [ b ɨ s ] | sit (imperative.2 p.sg.) | |
| | | | | [ p ɨ d ] | a traditional measure | |
| ə | a | | | [ k ə r ] | tax | |
| | | | | [ b ə s ] | bus | |
| | | | | [ p ə d ] | fall (imperative.2 p.sg.) | |
| a | | आ | ಆ | [ a ɖ s ə r ] | tender coconut | |
| | | | | [ r a dʒ a ] | king | |
| u | u | उ and ऊ | ಉ and ಊ | [ uːs ] | sugarcane | Back |
| | | | | [ p uː l ] | bridge | |
| o | o | ओ | ಒ | [ ts oː r] | thief | |
| | | | | [ d oː n ] | two | |
| ɔ | | ऑ | ಒ | [ ts ɔːr ] | thieves | |
| | | | | [ b ɔː l ] | ball | |

(Fadte et al., 2022)

Table 2: Oral vowels of Konkani

positions of the tongue in case of the nasal vowels. They have showed that the position of the tongue is generally lowered for Back vowels, fronted for Low vowels, and raised for Front vowels.
(Feng and Castelli, 1996) have presented work on the nasalisation of 11 French vowels. They show that the first two resonance frequencies are at about 300 and 1000 Hz.
(Carignan, 2014) is an acoustic study of three French oral-nasal vowel pairs /a/-/ã/, /ɛ/-/ɛ̃/, and /o/-/õ/. His study shows that the oral articulation of French nasal vowels is not arbitrary.

### 1.4. Objectives of the Present Study

As mentioned earlier, there are differences among Konkani scholars on the exact number of Vowel phonemes in the language. In the absence of more accurate descriptions of the Vowel system of the language and nasalisation of vowels, this study makes an attempt to target an important aspect of Konkani vowel phonemes namely, their nasalisation using acoustic analysis. For this study, we have taken into consideration the nine vowel phonemes mentioned in the Standard variety of the language which are the same as the ones mentioned in (Sardesai, 1986) and later cited in (Fadte et al., 2022).

This work is arranged into four sections. Section 1. - the Introduction section above highlights the linguistic features of the language and states the objective of the study. Section 2. discusses the methodology followed in the experiment. The results of the experiment are presented in section 3.. Section 4. concludes the paper with the scope for further studies.

### 1.5. Hypothesis

Given that nasalisation is phonemic in the language, each vowel phoneme of the language will have a nasal counterpart. In other words, the status of nasalisation as being phonemic in the language will become more explicit through this study.

## 2. Methodology

This section presents the details of the experimental work carried out and the methodology that was used for the experiment. (Fadte et al., 2022)'s methodology was followed for carrying out this experiment.

### 2.1. Recording Script

The recording script of this work was based on the Vowel phonemes mentioned in the classification provided by (Fadte et al., 2022). The Vowel phonemes in the script were arranged according to their classification which was established using the minimal and near-minimal pairs. A Phoneme is the smallest distinctive/contrastive unit in the sound system of a language. It is that unit of sound (a phone) that can distinguish one word from another in a particular language. The inventory of phonemes of a language is created using the minimal pairs (or near-minimal pairs in the absence of minimal pairs) of the language. Minimal pairs are pairs of words or phrases in a particular language that differ in only one phonological element and have distinct meanings. Near minimal pairs are pairs which have one or more additional differences elsewhere in the word besides the crucial position. Thus minimal pairs are an important tool that helps in establishing phonemes of a particular language. Pronunciation of phones is shown using square brackets whereas phonemes established using the minimal pairs are written in between slashes. To give an example, the Konkani words [na:k] 'nose' and [na:g] 'king cobra' differ only in the sounds [k] and [g]. Thus, the phones [k] and [g] in these words produce a difference in meaning. Using the [na:k] and [na:g] (minimal) pair we can now establish that the consonants [k] and [g] are phonemes of the language. These phonemes will therefore be written as /k/ and /g/. The recording script was created with Konkani sentences consisting of minimal pairs that aimed at establishing the vowel phonemes. A few examples of minimal pairs targeting vowel phonemes used in the recording script are provided in Table 3, and the entire script can be accessed from **here**.[1] The recording script consisted of 74 unique sentences, 37 for oral and 37 for nasal vowels, respectively. At least two different sets of minimal pairs were used for each vowel phoneme.

| Oral Vowel | | | | Nasal Vowel | | | |
|---|---|---|---|---|---|---|---|
| Konkani Examples | Vowel | IPA | Gloss | Konkani Examples | vowel | IPA | Gloss |
| शी | i | [ʃi:] | ugh' excl. | शीं | ĩ | [ʃĩ:] | cold' n/f.sg. |
| केस | e | [ke:s] | suit' f.sg. | केंस | ẽ | [kẽ:s] | a hair' m.sg. |
| वेत | ɛ | [vɛt] | cane' n.sg. | वेंत | ɛ̃ | [vɛt̃] | span' n.sg. |
| वय | ə | [və:j] | 'age'.neut.sg. | वंय | ə̃ | [və̃:j] | 'fence'.f.sg. |
| बाय | a | [ba:j] | 'girl'.neut.sg. | बांय | ã | [bã:j] | 'well'.f.sg. |
| खूट | u | [kʰu:t] | shortage' f.sg. | खूंट | ũ | [kʰũ:t] | stake' m.sg. |
| घोगो | ɔ | [gʰɔgɔ] | fall' m.sg. | घोंगो | ɔ̃ | [gʰɔ̃gɔ] | horn' m.sg. |

Table 3: Example of oral and nasal vowels in Konkani

### 2.2. Speakers' Detail

Three male and three female native speakers of Konkani were selected for this experiment. Speakers belonged to different geographical locations and spoke diverse regional dialects (details of these can be accessed from **here**[1]).This ensured that phone variability across regions was captured. All the speakers selected for the recording were literate.

### 2.3. Data Elicitation and Recording

The reading material consisting of sentences having minimal pairs was provided to the speakers as a printed copy.

---
[1] https://github.com/shashwatup9k/dhvani-konkani

The speakers were given some time to familiarise themselves with the meaning of the sentences. Then, they were instructed to read the sentences in the most natural way they could. Each sentence in the recording script was pronounced thrice by the speakers. This was done in order to capture any dialectal variation in the pronunciation of the target phonemes. It also helped to detect speaker-specific errors in phone production. The recording was performed in a closed room with less ambient noise. The audio was recorded using a Zoom-H6 recorder at a sampling rate of 48 kHz and was stored in non-lousy WAV format.

### 2.4. Annotation

Phoneme-level annotation of audio data was then carried out. Only the vowel phonemes which were to be used for analysis were annotated. A total of 1135 vowel phones were annotated in the data set. The frequency distribution of phones is presented in Figure 1.

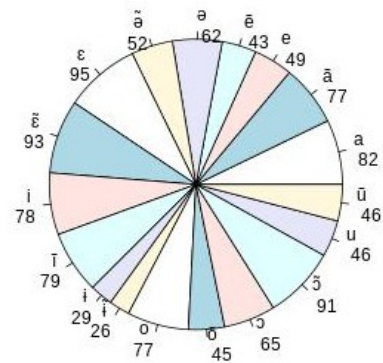

Figure 1: Frequency distribution of phones in dataset

Annotation was performed using the Praat software (Paul and Weenink, 1992). The start and end of a phone boundary was marked as perceived by the ear of the annotator and with the help of a spectrogram in the Praat tool. A sample of an audio signal, spectrogram and phoneme level annotation done in the Praat tool is shown in Figure 2.

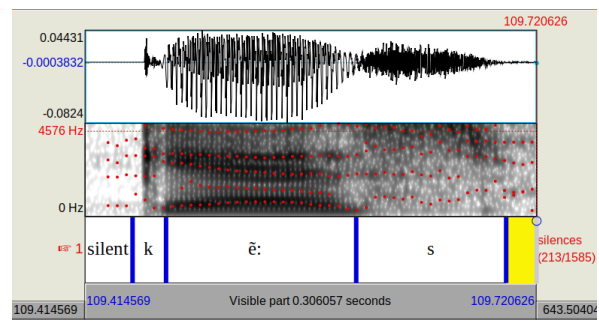

Figure 2: Annotations and spectrogram of phones

### 2.5. Formant Extraction

A Praat script was used to perform formant extraction on annotated data. This script extracts the formant details from the mid-temporal interval of the phoneme, which is stored in a text file. For formant extraction, values for speakers' frequency were set to standard values, i.e. 5 kHz for male and 5.5 kHz for female speakers. The data was stored in a text file and later converted to a CSV file for plotting results and analysis.

## 2.6. Data Verification

After formant details were extracted, they were plotted using a box plot for verifying visually any outliers that may have occurred due to wrong annotation. This step helped in identifying certain incorrect annotations, which were then corrected. As discussed in section 2.5., formant extraction was performed again, and boxplots were replotted. Box plots of the F1 and F2 formant for male speakers are shown in Figure 3. After the above corrections, a few outliers can still be seen.

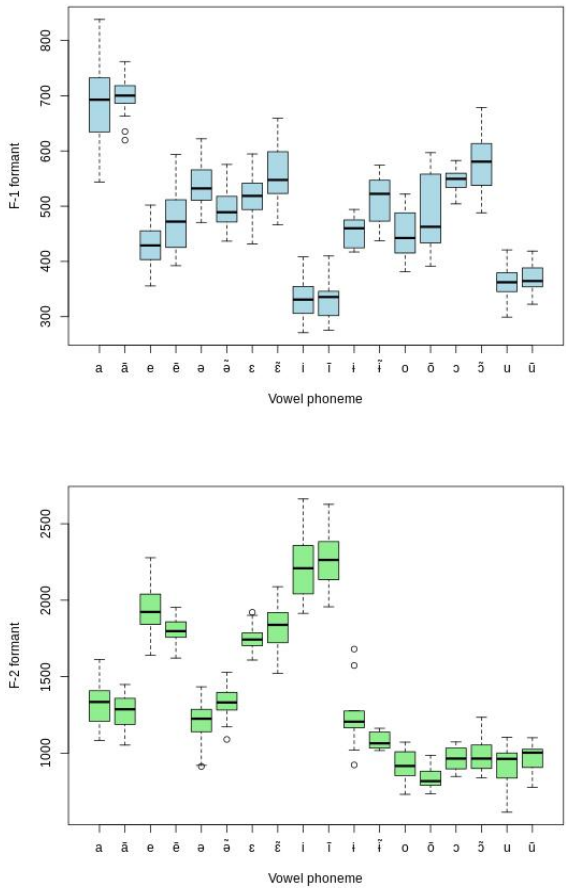

Figure 3: Box plot of F1 and F2 formants for oral and nasal vowels for male speakers

Final data verification was done with the help of a linguist who simultaneously listened to the phones and viewed the spectrogram to verify the label given to them. Errors in the phoneme production were noted down and are discussed in detail in section 2.7. below.

## 2.7. Substitution Analysis

The verified data showed some deviations from the expected production of the phonemes. Table 4 presents the phoneme substitution that occurred during the elicitation process. All these deviations were not used for the formant analysis. The close-mid central vowel [ɵ] with a frequency of 12 (see Table 4) occurring in place of the schwa ([ə]), is its allophone which occurs in the environment wherein it is followed by the open-mid back (rounded) vowel [ɔ] as in the words [ɡəʊɔ] 'bison' n.mas.sg, [bərɔ] 'good' adj.mas.sg.

| Phoneme | Substitution | Frequency | % of total substitution |
|---|---|---|---|
| i | None | 0 | 0.0 |
| ĩ | i | 28 | 12.4 |
| e | a | 1 | 0.4 |
| e | ɛ | 3 | 1.3 |
| ẽ | e | 6 | 2.7 |
| ẽ | ɛ | 12 | 5.3 |
| ẽ | ɛ̃ | 4 | 1.8 |
| ɛ | e | 6 | 2.7 |
| ɛ̃ | ɛ | 20 | 8.9 |
| ɨ | ə | 6 | 2.7 |
| ɨ | ə̃ | 3 | 1.3 |
| ɨ̃ | ə | 7 | 3.1 |
| ɨ̃ | ə̃ | 10 | 4.4 |
| ɨ̃ | ɵ | 1 | 0.4 |
| ə | ɵ | 12 | 5.3 |
| ə | ɨ | 4 | 1.8 |
| ə̃ | ɨ | 8 | 3.6 |
| ə̃ | ə | 6 | 2.7 |
| ə̃ | ã | 1 | 0.4 |
| ə̃ | o | 1 | 0.4 |
| a | ã | 8 | 3.6 |
| ã | a | 12 | 5.3 |
| u | ũ | 3 | 1.3 |
| ũ | u | 19 | 8.4 |
| o | ɔ | 12 | 0.9 |
| o | o | 8 | 3.1 |
| o | ɔ̃ | 2 | 0.9 |
| ɔ | ɔ̃ | 1 | 0.4 |
| ɔ̃ | ɔ | 32 | 14.2 |

Table 4: Substitution Analysis

## 3. Experimental Results and Analysis

A formant analysis of Vowel phonemes was performed as part of this study. R script was written with the use of the *phonR* package(McCloy, 2016) to plot the experimental results.

F1-F2 plots for oral and nasal vowels of male speakers are presented in Figure 4. A well-defined grouping of vowels in formant space is observed. From Figure 4, it is clearly seen that the Front oral vowels /i/, /e/ and /ɛ/ occupy non-intersecting space in the formant chart. In the same figure, we can see that the three extreme vowels (/i/, /a/, and /u/) occupy three corners in formant space. Other vowels also do not have many intersections in formant space. F1-F2 plots for oral and nasal vowels of female speakers are presented in Figure 5, which have similar features as seen in the male formant chart. A comparative chart for additional details can be accessed **here**[1]. Apart from the formant charts, we have also listed the average values of F1, F2, and F3 formants for male and female speakers. These are presented in Table 5. The average values for the F1 and F2 for oral vowels are similar to those reported by (Fadte et al., 2022). Since no previous work was reported for the nasal values, formant values provided in Table 5 may be considered as the first reporting of such work.

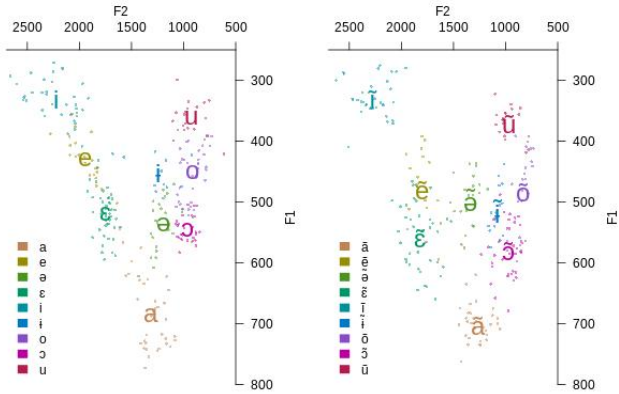

Figure 4: Formant chart for oral and nasal vowels for male speakers

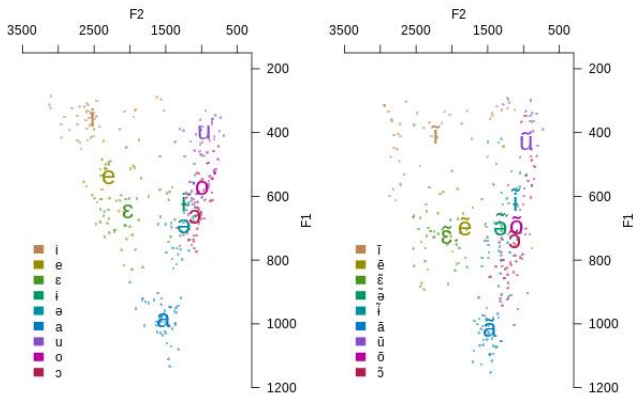

Figure 5: Formant chart for oral and nasal vowels for female speakers

## 4. Conclusion and Future Work

This work provides a comparative study of the Konkani Vowel phonemes (i.e. oral and nasal vowels). The results have shown that all oral vowel sounds in Konkani can be nasalised. It is observed that the different vowels in the formant chart are in their expected position as per (Fadte et al., 2022) vowel classification. It is also seen that nasalisation changes the F1-F2 values for the vowel phonemes.

| Phoneme | female | | | male | | |
|---|---|---|---|---|---|---|
| | F1 | F2 | F3 | F1 | F2 | F3 |
| i | 353 | 2518 | 3055 | 331 | 2229 | 2730 |
| e | 535 | 2240 | 2773 | 428 | 1943 | 2470 |
| ɛ | 641 | 2518 | 3055 | 331 | 1749 | 2450 |
| ɨ | 636 | 1339 | 2978 | 453 | 1245 | 2548 |
| ə | 690 | 1224 | 3081 | 537 | 1193 | 2508 |
| a | 982 | 2518 | 3055 | 685 | 1319 | 2446 |
| u | 417 | 972 | 2904 | 362 | 930 | 2464 |
| o | 574 | 990 | 3072 | 450 | 914 | 2585 |
| ɔ | 670 | 1089 | 2964 | 544 | 966 | 2452 |
| ĩ | 393 | 2204 | 3049 | 328 | 2272 | 2747 |
| ẽ | 673 | 1861 | 2699 | 477 | 1800 | 2512 |
| ɛ̃ | 708 | 2057 | 2724 | 556 | 1818 | 2434 |
| ɨ̃ | 630 | 1150 | 2890 | 514 | 1082 | 2515 |
| ə̃ | 688 | 1326 | 2914 | 496 | 1341 | 2526 |
| ã | 974 | 1461 | 2737 | 699 | 1271 | 2413 |
| ũ | 402 | 999 | 2749 | 368 | 969 | 2442 |
| õ | 691 | 1103 | 2941 | 480 | 838 | 2644 |
| ɔ̃ | 726 | 1138 | 2985 | 576 | 977 | 2497 |

Table 5: Average F1, F2, and F3 values for Vowel phonemes.

The average F1, F2, and F3 values for nasal vowels are reported for the first time through experimentation. This work can be helpful for the linguistic study of vowels and speech synthesis systems specific to Konkani language. Although oral and nasal studies have been presented, other phones and combinations of phones, like consonants, diphthongs have not been explored in this work or rather there have not been acoustic studies done related to the properties of such phones in Konkani language. We wish to explore these in our future work.

## Acknowledgements

Atul Kr. Ojha would like to acknowledge the support of the Science Foundation Ireland (SFI) as part of Grant Number SFI/12/RC/2289_P2, Insight SFI Centre for Data Analytics.